\documentclass[letterpaper, 10 pt, journal, twoside]{IEEEtran}
\usepackage{amsmath,amsfonts}
\usepackage{algorithmic}
\usepackage{array}
\usepackage[caption=false,font=normalsize,labelfont=sf,textfont=sf]{subfig}
\usepackage{textcomp}
\usepackage{stfloats}
\usepackage{url}
\usepackage{verbatim}
\usepackage{graphicx}
\usepackage{cite}
\usepackage[ruled,vlined,linesnumbered]{algorithm2e}
\SetCommentSty{emph}
\usepackage{multirow}
\usepackage{graphicx} 
\usepackage{url}
\usepackage{xcolor}
\usepackage{amssymb,amsmath,comment}
\usepackage{amsthm}
\usepackage{amsfonts}
\usepackage[capitalize]{cleveref}
\usepackage{booktabs}
\Crefname{equation}{Eq.}{Eqs.}
\Crefname{figure}{Fig.}{Figs.}
\Crefname{tabular}{Tab.}{Tabs.}

\newcommand{\mymethod}{SMART\xspace}
\newcommand{\repolink}{\url{https://github.com/smart-mapf/smart}}

\begin{document}

\title{Advancing MAPF Toward the Real World: \\A Scalable Multi-Agent Realistic Testbed (\mymethod)}


\author{
{Jingtian Yan$^1$,
Zhifei Li$^1$,
William Kang$^{1}$,
Kevin Zheng$^{2}$,
Yulun Zhang$^1$,
Zhe Chen\textsuperscript{\rm 2},
Yue Zhang\textsuperscript{\rm 2},\\
Daniel Harabor\textsuperscript{\rm 2},
Stephen F. Smith$^1$,
Jiaoyang Li$^1$}
\thanks{$^{1}$J. Yan, Z. Li, W. Kang, Y. Zhang, S. Smith, and J. Li are with Robotics Institute, Carnegie Mellon University, Pittsburgh, PA 15213, USA {(email: \{jingtianyan, zhifeil, williamkang, yulunzhang\}@cmu.edu, ssmith@andrew.cmu.edu, jiaoyangli@cmu.edu)}.}
\thanks{$^{2}$K. Zheng, Z. Chen, Y. Zhang, and D. Harabor are with Monash University, Clayton, VIC 3800, Australia. {(email: \{kevin.zheng, zhe.chen, yue.zhang, daniel.harabor\}@monash.edu)}.}
\thanks{Digital Object Identifier (DOI): see top of this page.}
}

\markboth{Journal of \LaTeX\ Class Files,~Vol.~14, No.~8, August~2021}%
{Shell \MakeLowercase{\textit{et al.}}: A Sample Article Using IEEEtran.cls for IEEE Journals}


\maketitle

\begin{abstract}
We present \textbf{S}calable \textbf{M}ulti-\textbf{A}gent \textbf{R}ealistic \textbf{T}estbed (\mymethod), a realistic and efficient software tool for evaluating Multi-Agent Path Finding (MAPF) algorithms.
MAPF focuses on planning collision-free paths for a group of robots. 
While state-of-the-art MAPF planners can plan paths for hundreds of robots in seconds, they often rely on simplified robot models, making their real-world performance unclear.
Researchers typically lack access to hundreds of physical robots in laboratory settings to evaluate the algorithms.
Meanwhile, industrial professionals who lack expertise in MAPF require an easy-to-use simulator to efficiently test and understand the performance of MAPF planners in their specific settings.
\mymethod fills this gap with several advantages:
(1) \mymethod uses physics-engine-based simulators to create realistic simulation environments, accounting for complex real-world factors such as robot kinodynamics and execution uncertainties,
(2) \mymethod uses an execution monitor framework based on the Action Dependency Graph, facilitating seamless integration with various MAPF planners and robot models, and
(3) \mymethod scales to thousands of robots.
The code is publicly available at \repolink \  with an online service available at \url{https://smart-mapf.github.io/demo/}.
\end{abstract}


\section{Introduction}
The Multi-Agent Path Finding (MAPF) problem~\cite{Stern2019benchmark} focuses on planning collision-free paths for a large group of robots within a known environment.
This problem finds various real-world applications~\cite{ma2017feasibility,honig2019warehouse,Ho2022uav}, with warehouse automation being a prominent example, where hundreds of robots must be coordinated to perform daily operations.
In recent years, MAPF has grown into a rapidly expanding research area~\cite{ma2022graphbased,wang2025wherepaths}.
State-of-the-art MAPF planners can plan paths for hundreds of robots within seconds~\cite{ma2019searching,ZhangAIJ22,li2022mapf}.
However, these algorithms make several simplifying assumptions.
First, they rely on simplified robot models that ignore kinodynamic constraints while planning the robots' paths.
Second, they assume that robots can execute these paths perfectly, without accounting for uncertainties introduced by real-world factors.
Therefore, while researchers have shown that MAPF planners have a wide range of applications, their performance in more realistic settings is unclear.

\begin{figure}
    \centering
    \includegraphics[width=\linewidth]{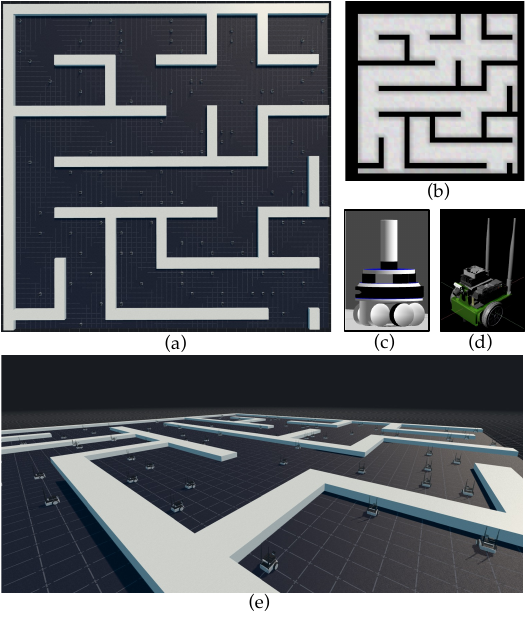}
    \caption{(a) A simulation environment in \mymethod.
    (b) The original 2-D grid map from the MovingAI benchmark.
    (c, d) Robot models used in the environment.
    (e) A 3D view of the environment.}
    \label{fig:intro_idea}
\end{figure}

MAPF researchers in both academia and industry require an efficient, realistic, and scalable software tool to evaluate MAPF planners in realistic settings.
In this paper, we use the term \emph{scalability} to refer to runtime capacity in terms of the maximum number of robots that the testbed can support in simulation.
In academia, deploying hundreds of physical robots in a large environment to evaluate MAPF planners is impractical. 
Meanwhile, developing software that can evaluate large groups of robots is a non-trivial task for MAPF researchers.
Researchers must create high-fidelity simulation environments and robot models to capture realistic factors, as well as develop essential software components such as controllers and execution frameworks to execute their plans.
Moreover, since many researchers come from the AI community, learning such robotic simulation tools can require significant effort.
While some software tools~\cite{schaefer2023benchmark,heuer2024benchmarking} have been developed to evaluate MAPF planners in realistic settings, they do not scale to hundreds of robots and require significant engineering efforts to integrate new MAPF planners and maps.
In addition, industrial professionals possess realistic simulators for their applications of interest. However, they may not have the resources and expertise to implement state-of-the-art MAPF planners in their simulators. By having a realistic simulator that can easily interface with existing implementations of MAPF planners, industrial professionals can more easily understand and select a MAPF planner that fits their demands.

In this paper, we present \mymethod, a \textbf{S}calable \textbf{M}ulti-\textbf{A}gent \textbf{R}ealistic \textbf{T}estbed, a software tool used to evaluate MAPF planners in realistic scenarios.
\mymethod comprises three main components: (1) a physics-engine-based simulator, (2) an execution monitoring (EM) server, and (3) robot-specific executors.
The simulator generates realistic simulation environments based on user-defined configurations.
\Cref{fig:intro_idea} shows an example environment in \mymethod created from a 2-D grid map from the MovingAI benchmark~\cite{Stern2019benchmark}.
The EM server parses the user-provided MAPF paths, tracks the execution progress, and communicates with the executors.
It leverages the Action Dependency Graph (ADG)~\cite{honig2019warehouse}, a MAPF execution framework, to ensure robust execution of the paths generated by MAPF planners using various simplified robot models.
Each robot is assigned an individual executor that receives and executes actions from the server.
\mymethod incorporates key real-world factors, including robot kinodynamics, collision dynamics, communication delays, execution imperfections, and temporal uncertainty during execution.

Our contributions in this paper are as follows:
1. We present \mymethod, an open-source software tool for evaluating MAPF planners in realistic scenarios. \mymethod directly accepts MAPF plans and problem instances from existing benchmarks.
2. We integrate \mymethod with two simulators, ARGoS3~\cite{pinciroli2012argos} and Isaac Sim~\cite{nvidiaisaacsim}, and empirically evaluate its scalability and replicability. \mymethod can scale to thousands of robots while maintaining consistent and reproducible results, whereas prior tools fail to handle a hundred robots.
3. We develop an online interface with interactive visualization, allowing users to easily monitor execution progress and access detailed statistical data during simulation.
4. We validate \mymethod on real-world mobile robots, demonstrating its robustness and effectiveness in physical environments.

\begin{figure}
    \centering
    \includegraphics[width=\linewidth]{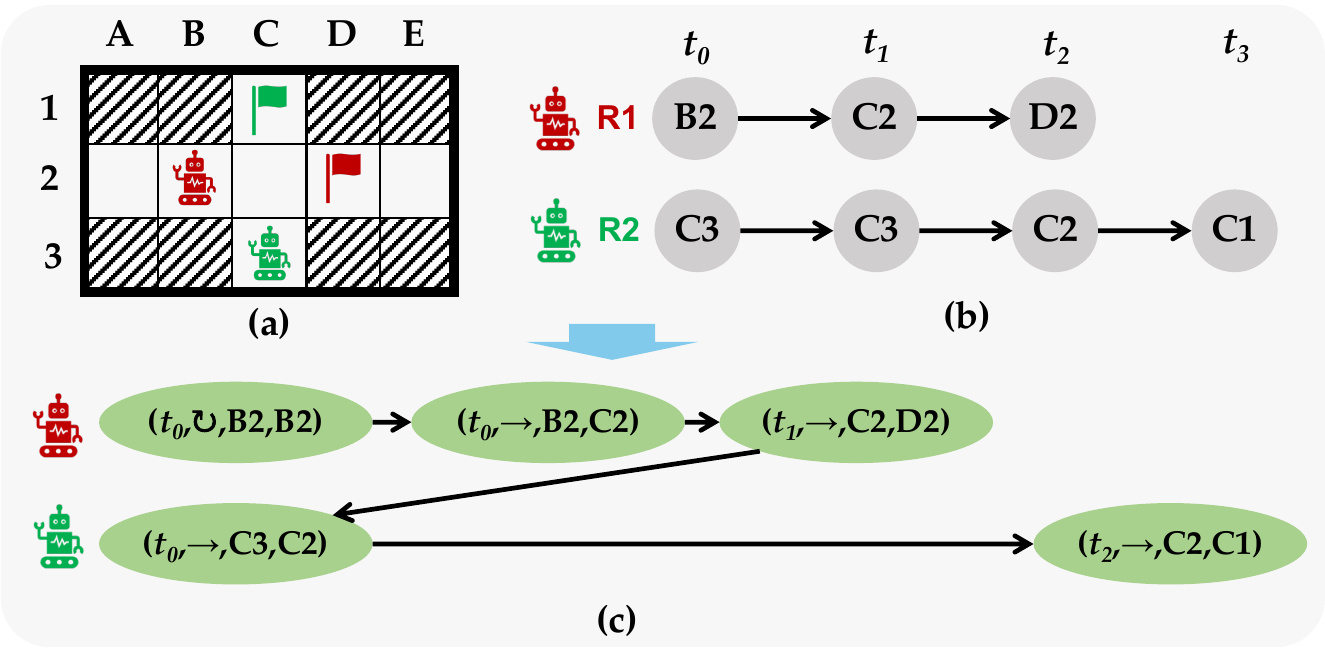}
    \caption{(a) MAPF problem. (b) Associated MAPF plan. (c) ADG from the MAPF plan. Green ovals are actions, with the first element as the reference time from the plan, the second as the action type (e.g., rotation or forward movement), and the last two as start and goal locations. Horizontal edges are Type-1, and vertical edges are Type-2. All robots initially face north.}
    \label{fig:example_adg}
\end{figure}

\section{Background}
In this section, we first introduce the MAPF problem and related work. Next, we discuss existing execution frameworks for executing MAPF plans. Finally, we review benchmarks used to evaluate MAPF planners.

\subsection{MAPF Problem and MAPF Plan}
A \emph{MAPF instance}~\cite{Stern2019benchmark} consists of an undirected graph $\mathcal{G}_M = (\mathcal{V}_M, \mathcal{E}_M)$ constructed from a 2-D grid map and a set of $I$ robots $\mathcal{R} = \{r_1, \dots, r_I\}$ with their corresponding start and goal locations.
Given a MAPF instance, a MAPF planner aims to find a \emph{MAPF plan}, consisting of collision-free paths for all robots from their start to goal locations. Standard MAPF uses a simplified robot model by discretizing the time into timesteps and assuming that, at each timestep, every robot either moves to an adjacent vertex or stays at its current vertex. Two robots collide if they are at the same vertex or swap vertices at the same timestep.
An example MAPF instance and plan on a 2-D grid graph are shown in \Cref{fig:example_adg} (a) and (b), respectively.
The typical objective of MAPF is to minimize the sum-of-cost, defined as the sum of travel time.

Prior research has proposed many MAPF planners, most of which were implemented and tested on 2-D grid maps.
Some methods focus on finding solutions with optimal or bounded-suboptimal guarantees~\cite{sharon2015conflict,li2021eecbs}, while others focus on scalability, solving MAPF problems involving thousands of robots~\cite{okumura2022priority,okumura2023lacam}.
Additionally, some methods incorporate complex real-world factors, such as kinodynamics and unexpected delays, into the planning process~\cite{cohen2019optimal,atzmon2020robust,yan2024PSB,yan2025multi}.
They usually employ a two-level planner, where the high level employs MAPF-based methods to resolve collisions among robots, while the low level plans for each robot considering real-world factors.

\subsection{Executing MAPF Plans}
Given a MAPF plan, robots can remain collision-free if they strictly follow it spatially and temporally.
While existing path-tracking techniques allow for reasonable spatial accuracy, precise temporal adherence is more challenging~\cite{vcap2016provably}, especially since MAPF plans are typically generated using imperfect robot models.
Thus, we need techniques to execute MAPF plans safely given the uncertainty in execution time.
One naive way is synchronized execution, where robots need to wait for all robots to finish their current actions before starting the next. This ensures strict compliance with the MAPF plan but significantly reduces efficiency.
Alternatively, one can replan whenever a robot is delayed, but this is computationally expensive and thus does not scale well.
Therefore, executing MAPF plans using an execution framework is preferred~\cite{vcap2016provably}. The state-of-the-art execution framework is Action Dependency Graph (ADG)~\cite{honig2019warehouse}, which enforces passing orders defined by MAPF plans while accounting for delays.
ADG can execute plans generated by MAPF planners of various robot models~\cite{varambally2022mapf}.

An ADG, as shown in~\cref{fig:example_adg} (c), is a directed graph $\mathcal{G} = (\mathcal{V}, \mathcal{E}_1, \mathcal{E}_2)$ that encodes dependencies between actions from a MAPF plan.
The vertex set $\mathcal{V} = \{v_i^k : k \in [1, n^i], r_i \in \mathcal{R}\}$ represents all actions performed by each robot $r_i$, where each vertex $v_i^k$ corresponds to the $k$-th action of $r_i$, and $n^i$ is its number of actions.
The edge set $\mathcal{E}_1$ includes \emph{Type-1 edges}, which enforce the sequential execution of actions by each robot. For a robot $r_i$, a Type-1 edge $(v_i^k, v_i^{k+1})$ ensures that action $v_i^k$ is completed before action $v_i^{k+1}$ starts.
The edge set $\mathcal{E}_2$ includes \emph{Type-2 edges}, which enforce the inter-robot passing order at shared vertices.
For any two vertices $v_i^k$ and $v_j^s$ from $r_i$ and $r_j$, a Type-2 edge $(v_i^k, v_j^s)$ ensures that $v_i^k$ must be completed before $v_j^s$ can start.
During execution, each action is performed according to these dependencies. The detailed execution procedure is described in~\cref{method:server}.

Although modern MAPF solvers have become significantly faster, execution frameworks such as ADG remain essential due to inevitable real-world uncertainties. Some of them, such as communication delays, cannot be eliminated through faster planning alone. Moreover, by enforcing only necessary ordering constraints rather than strict timestep synchronization, ADG enables robots to execute multiple consecutive actions without stopping, supporting velocity optimization and smoother execution rather than requiring stop-and-go behavior at every timestep.


\begin{table*}[t]
\centering
\caption{Comparison of SMART with existing multi-robot frameworks and evaluation tools. Interactive execution diagnostics refer to inspection of execution states during runtime, and benchmark support refers to compatibility with standard MAPF benchmarks.}
\label{tab:comparison}
\begin{tabular}{lcccccc}
\toprule
\textbf{Tool} & \textbf{Max Robots} & \textbf{MAPF Planner} & \textbf{General Execution} & \textbf{Interactive Execution} & \textbf{Benchmark} & \textbf{Primary} \\
 & \textbf{(Tested)} & \textbf{Integration} & \textbf{Framework} & \textbf{Diagnostics} & \textbf{Support} & \textbf{Purpose} \\
\midrule
Open-RMF~\cite{openrmf} & $\sim$100 & Manual & Yes & No & No & Fleet deployment \\
MRP-Bench~\cite{schaefer2023benchmark} & $<$10 & Manual & No & No & Partial & Small-scale MAPF evaluation \\
REMROC~\cite{heuer2024benchmarking} & $<$50 & Manual & No & No & No & Human-aware evaluation \\
{SMART (ours)} & {2000} & Yes & Yes & {Yes} & {Yes} & {MAPF research} \\
\bottomrule
\end{tabular}
\end{table*}

\subsection{Evaluating MAPF Plans}
To evaluate MAPF planners, researchers typically generate a plan for a given MAPF instance and then execute it.
While many MAPF benchmarks with 2-D grid maps are proposed~\cite{Stern2019benchmark,gebser2018experimenting,sturtevant2012benchmark,Qian2024Benchmark}, most prior MAPF works are evaluated on them using simplified robot models that assume no delays and discretized time and space.
Some benchmarks have attempted to include realistic elements such as acceleration and rotation~\cite{mohanty2020flatlandrl,chan2024league}. However, their models are still far from realistic, and they typically require the same robot models for planning and execution phases.

Given the difficulty of modeling all real-world factors analytically, physics-engine-based simulators offer a promising approach for more accurate evaluation.
Frameworks like Open-RMF~\cite{openrmf} provide middleware for deploying and coordinating robot fleets in production environments, handling task allocation, traffic management, and building infrastructure integration.
However, such operational deployment frameworks are not specialized for MAPF evaluation. While centralized MAPF planners can be integrated, using them to benchmark and stress-test MAPF planners at large scale typically requires additional engineering effort beyond their core scope.
To address the need for MAPF-specific evaluation, several simulation-based tools have been developed.
MRP-Bench~\cite{schaefer2023benchmark} leverages a high-fidelity, physics-based simulator Gazebo~\cite{koenig2004design} to simulate multi-robot systems. It integrates low-level controllers to account for robots’ kinodynamic constraints and potential delays. REMROC~\cite{heuer2024benchmarking} also uses Gazebo to realistically evaluate MAPF plans, with a particular focus on human-shared environments.
However, as shown in~\Cref{tab:comparison}, these existing tools have two main limitations. First, they often require significant engineering effort to integrate user-defined maps and new MAPF planners, limiting their flexibility for different scenarios.
In particular, they lack a general execution framework, making it difficult to directly evaluate MAPF planners across different robot models assumed during planning.
Second, they have limited scalability: the largest number of robots tested in previous work is less than 100 robots. Most of these works rely on Gazebo, whose high computational overhead, resulting from modeling many factors not relevant to MAPF, significantly constrains scalability.
MRP-Bench, for example, does not scale to more than 10 robots.

\begin{figure*} 
    \centering
    \includegraphics[width=\linewidth]{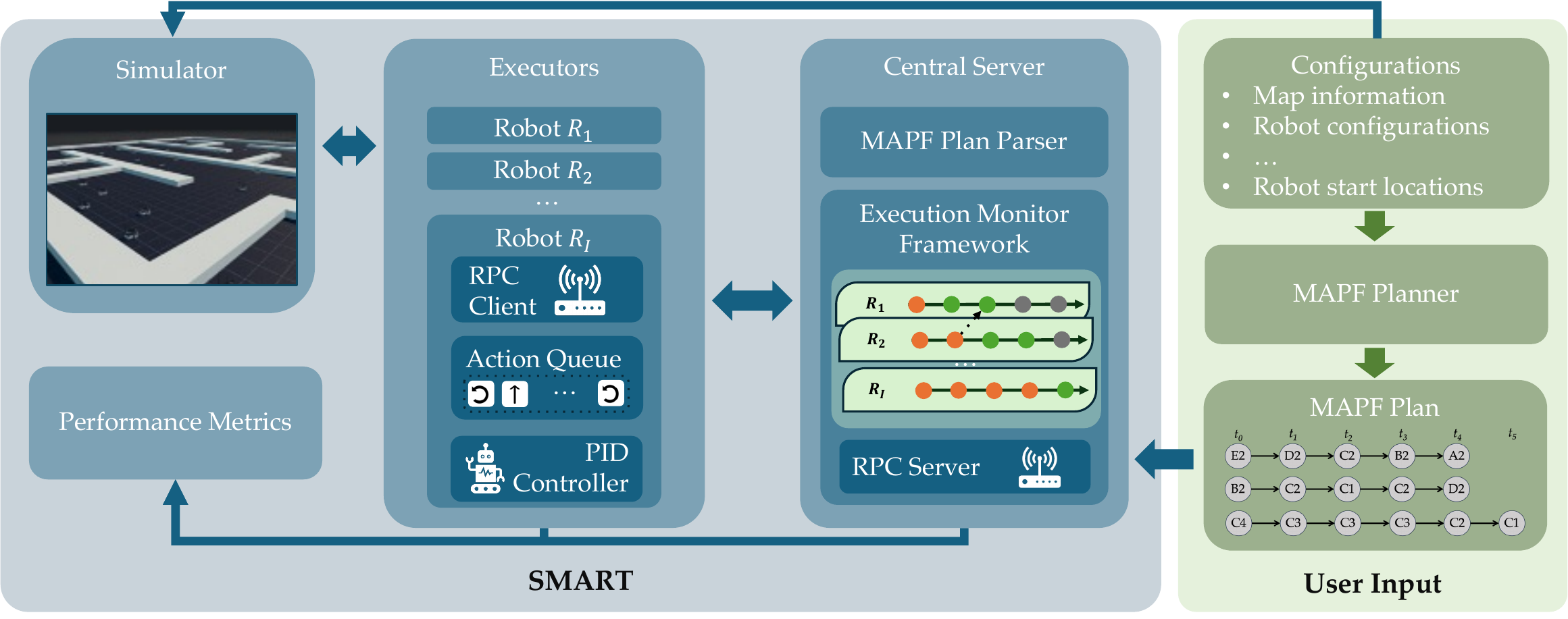}
    \caption{System overview of \mymethod.}
    \label{fig:system_overview}
\end{figure*}

\section{\mymethod Architecture}
\Cref{fig:system_overview} shows an overview of \mymethod. It comprises three main modules: a simulator, an execution monitoring (EM) server, and executors. All modules are implemented in C++.
The simulator generates simulation environments based on the user-provided configurations.
The EM server parses the MAPF plan and monitors the execution status of each action throughout the simulation. When an action is ready for execution, the server sends the action to the corresponding robot via the network.
The executor of each robot receives actions from the server, executes them using a controller, and synchronizes the execution status with the server accordingly.

\subsection{Simulator}
Given a user-provided configuration, the simulator creates a realistic simulation environment.
The configuration specifies obstacle locations and shapes, robot start locations, and robot configurations (e.g., velocity and acceleration limits).
To support the widely used MovingAI benchmark~\cite{Stern2019benchmark}, \mymethod can convert its 2-D grid maps and corresponding MAPF instances into the configuration required by the simulator.
During execution, the simulator uses a physics engine to update the states of the robots, including their poses, dynamics, and actuators, based on the received control commands from the executors. 
The simulator provides real-time robot state information (e.g., current pose) to other system components.
To accommodate diverse user requirements and balance scalability with physical realism, SMART supports two simulation platforms: ARGoS3~\cite{pinciroli2012argos} for efficient large-scale experiments and Isaac Sim~\cite{nvidiaisaacsim} for higher-fidelity simulation.

\subsubsection{ARGoS3}
ARGoS3 is a multi-robot simulator with excellent scalability, advanced physics engines, and support for custom robot and environment models, making it suitable for our tasks.
While users can specify any robot model, we use the foot-bot (shown in~\cref{fig:intro_idea} (c)), a differential-drive robot, in our experiments. It closely resembles robots used in real-world warehouses and can move forward and rotate in place.

\subsubsection{Isaac Sim}
Isaac Sim is a high-fidelity robotics simulator developed by NVIDIA, featuring GPU-accelerated physics, photorealistic rendering, and seamless integration with ROS. It is well-suited for sim-to-real transfer and physically grounded perception and control tasks.
We use the JetBot robot (shown in~\cref{fig:intro_idea} (d)), a differential-drive robot commonly used in research and education.

\begin{figure}
    \centering
    \includegraphics[width=1.0\linewidth]{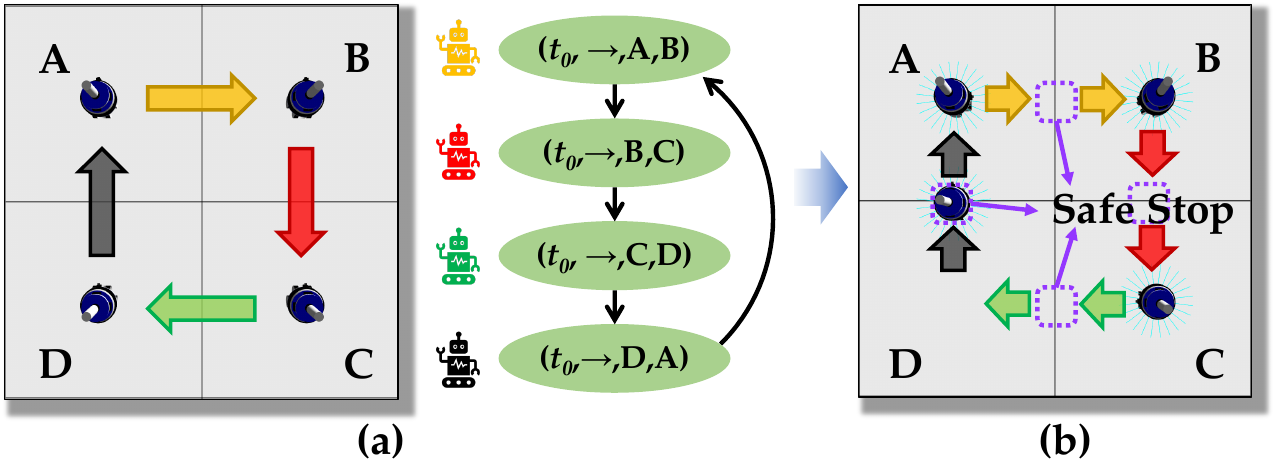}
    \caption{(a) Cycle in the ADG causing all robots to be stalled. (b) Adding safe stops between vertices to make the ADG acyclic. Robot diameter must be at most half the cell size to prevent collisions at safe stops between vertices.}
    \label{fig:adg_robust}
\end{figure}

\subsection{Execution Monitoring (EM) Server} \label{method:server}
The EM server parses the MAPF plan and encodes its passing order into an ADG. It then manages the execution of the MAPF plan and communicates with the executors.

\paragraph{Server Initialization}
During initialization, we use the MAPF plan parser to convert it into a sequence of actions and then construct an ADG based on these actions.
We define a lightweight, planner-agnostic standardized format to store the MAPF plan as a sequence of timed locations for each robot $r_i$, i.e., $((x_i^0, y_i^0, t_i^0),... , (x_i^T, y_i^T, t_i^T))$. This simple representation can be easily derived from any MAPF-generated path, making the parser compatible with any valid MAPF plan regardless of the underlying robot model or planner used to generate them. This design choice allows \mymethod to work seamlessly with planners using varying simplified robot models without requiring modifications to the planners themselves.
The parser first converts the path of each robot into a sequence of actions.
Specifically, for each transition between consecutive vertices, the parser infers the required robot orientation. If the orientation differs from the current orientation, an in-place rotation action is added. Subsequently, a forward movement action is added to move the robot to the next vertex.
The ADG is then constructed based on this sequence of actions.
ADG requires the MAPF plan to be cycle-conflict-free, i.e., it prohibits cyclic location exchanges among a set of robots within the same timestep. This is because ADG only allows movements into empty space to ensure delay safety, which cannot be satisfied under a cycle conflict~\cite{honig2019warehouse}.
However, ensuring this is not standard in most MAPF models.
As a result, if we directly build an ADG based on a plan with cycle conflicts, we may create deadlocks, such as the one illustrated in~\cref{fig:adg_robust} (a). While these robots are planned to rotate simultaneously in a cycle, with ADG, they will end up waiting indefinitely for others to complete their next actions.
To accommodate a broader range of planners, motivated by~\cite{honig2016multi}, we add an extra vertex between consecutive vertices in \mymethod as shown in~\cref{fig:adg_robust} (b), making the ADG compatible with these MAPF models.

\paragraph{Execution Management}
During execution, an execution monitor framework manages the progress of execution and communicates with robot executors.
This framework follows the method described in ADG~\cite{honig2019warehouse} to track and manage execution.
Each vertex in the ADG represents an action and can have one of three statuses: \emph{staged}, \emph{enqueued}, or \emph{finished}. 
At first, all vertices are \emph{staged}. 
A vertex transitions to \emph{enqueued} if (1) it has no preceding vertices, or (2) its preceding vertex connected by a Type-1 edge is \emph{enqueued} or \emph{finished}, and all its preceding vertices connected by Type-2 edges are \emph{finished}. Actions associated with \emph{enqueued} vertices are then sent to the executors for execution.
A vertex is marked as \emph{finished} when the executor confirms the completion of its associated action with the server. 
Communication with executors is managed by a Remote Procedure Call (RPC) server built with rpclib.\footnote{Available at \url{http://rpclib.net}.}

\subsection{Executor}
Each robot is launched with an individual executor. As shown in~\cref{fig:system_overview}, each executor consists of an RPC client, an action queue, and a controller. The RPC client communicates with the server by receiving \emph{enqueued} actions and sending confirmations for \emph{finished} actions. The action queue stores \emph{enqueued} actions and executes them in first-in-first-out order. A Proportional Integral Derivative (PID) controller executes these actions by sending control commands to the simulator, and a confirmation is sent to the server upon completion.
To allow robots to operate at higher speeds, consecutive actions of the same type are merged into one action in the queue.
For example, if a robot receives three consecutive forward movement actions, the executor merges them into a single longer forward action, reducing unnecessary stop-and-go behavior.
This distributed executor design enables parallel operation, as each robot handles the above-mentioned operations independently, improving scalability.

\subsection{Performance Metrics}
\mymethod provides several built-in performance metrics, including 
(1) \emph{Average execution time (AET):}
The sum of the simulation times required by all robots to complete their assigned paths, divided by the number of robots, 
(2) \emph{Maximum execution time:}
The longest simulation time taken by any single robot to complete its path, and
(3) \emph{Robot state over time:} The state of each robot at different times, where the state includes the size of its action queue, its current position, and its current wheel velocity.
Additionally, the open-source nature of our code allows researchers to easily derive custom metrics from \mymethod as needed.
\mymethod also provides visualization for the MAPF plan executions, allowing researchers to better understand the behaviors of the robots.

\section{System Usage}
Users can evaluate their solutions using \mymethod in two ways: by uploading their MAPF problem instance and plan through a web interface or by deploying \mymethod locally.

\begin{figure}[t]
    \centering
    \includegraphics[width=1.0\linewidth]{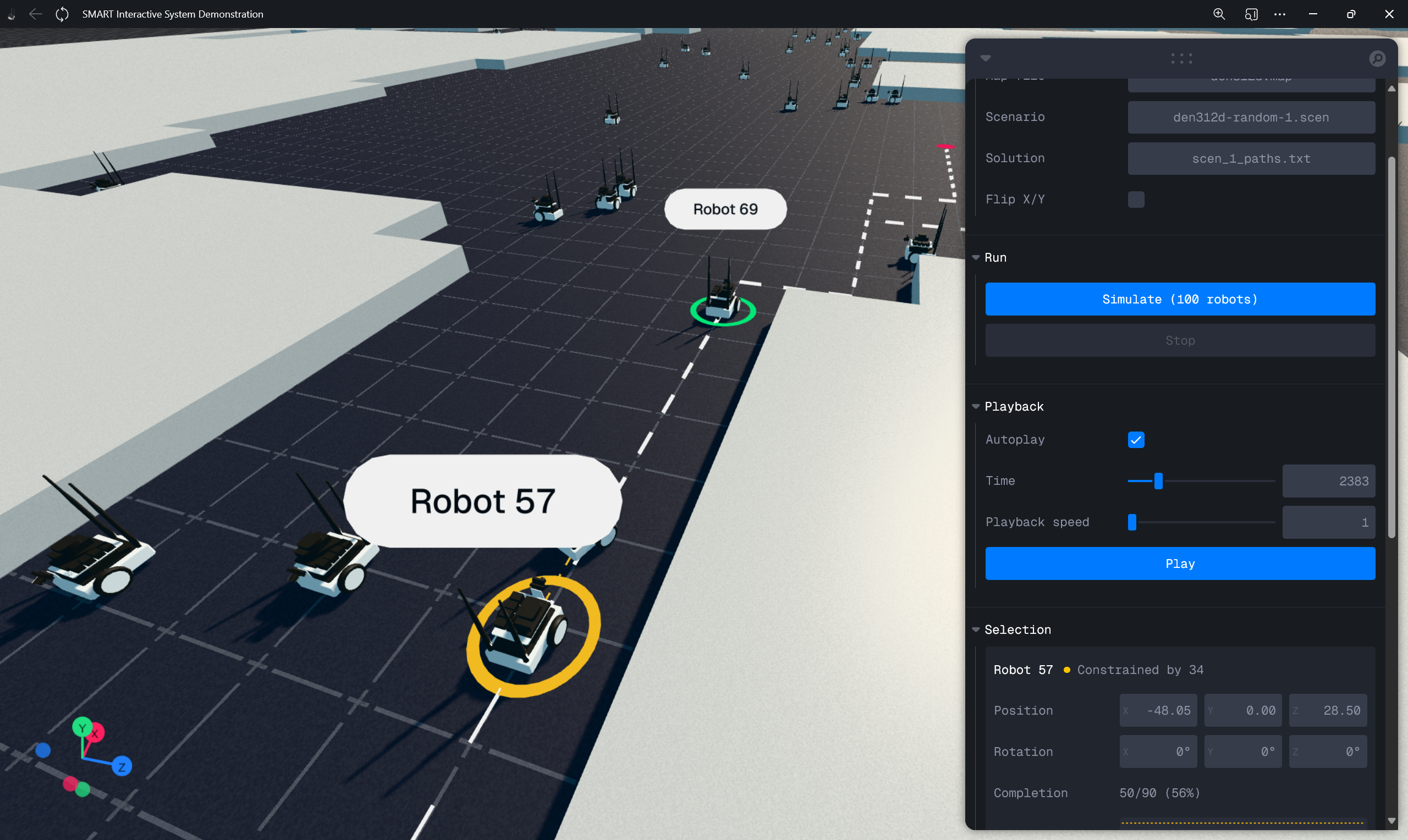}
    \caption{The \mymethod web interface visualizing a multi-robot simulation on a grid map. In the viewport, users interact with individual robots to inspect paths and inter-robot constraints. On the side panel, users can configure the environment, robot kinematic limits, control playback, and see execution details.}
    \label{fig:web_ui}
\end{figure}

\paragraph{\mymethod as a Service}
\mymethod, with ARGoS3 and the 3D visualizer, is directly available as software-as-a-service.
Specifically, we provide an intuitive web app for running and inspecting MAPF simulations. It features a 3D visualizer built with React and Three.js, allowing researchers to better understand how algorithms perform on physical robots (\Cref{fig:web_ui}). Users configure the environment by specifying a map and scenario file in the MovingAI Benchmark format.
Within the online interface, after users upload the files and select the desired kinodynamic settings for the robots, users
can initiate the simulation with a single click.
After starting the simulation, they can control the simulation time step and adjust the playback speed. Users may select robots to view full paths, execution status, and Type-2 edges between robots.

\paragraph{Installing \mymethod Locally}
Since \mymethod provides flexible interfaces for customization, researchers may choose to compile it locally for advanced use cases, such as running it in headless mode, integrating it into custom pipelines, or interfacing with real robots. As an open-source tool, \mymethod includes documentation and examples to support these advanced configurations.

\section{System Evaluation}
We test \mymethod on two machines. All experiments are conducted on machine (1), a high-performance server with a 64-core AMD 7980X processor, 256 GB of memory, and RTX 3090Ti GPU. In \Cref{subsec:usage}, we additionally test \mymethod on machine (2), a low-end laptop with a 4-core Intel i7-6700HQ processor and 16 GB of memory,  to demonstrate \mymethod{}’s performance on resource-constrained hardware. Both machines are equipped with Ubuntu 20.04.

\begin{figure}
    \centering
    \includegraphics[width=1.\linewidth]{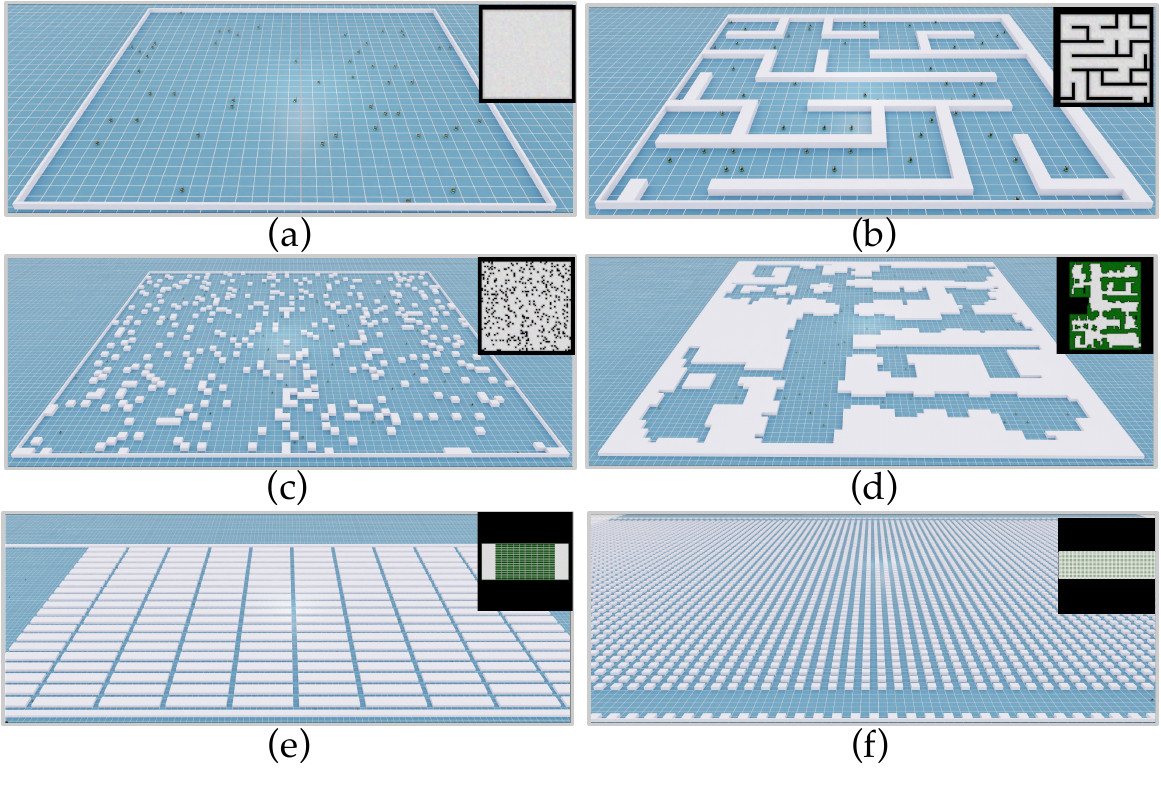}
    \caption{Simulation environments. (a) \texttt{empty} (b) \texttt{maze} (c) \texttt{random} (d) \texttt{game} (e) \texttt{warehouse} (f) \texttt{sortation-center}.}
    \label{fig:envs}
\end{figure}

As shown in~\Cref{fig:envs}, we evaluate \mymethod in six environments converted from 
\texttt{empty} (empty-32-32, size: 32$\times$32), \texttt{maze} (maze-32-32-4, size: 32$\times$32), \texttt{random} (random-64-64-10, size: 64$\times$64), \texttt{game} (den312d, size: 65$\times$81), and
\texttt{warehouse} (warehouse-10-20-10-2-1, size: 161$\times$63) from the MovingAI benchmark~\cite{Stern2019benchmark} and \texttt{sortation-center} (size: $500\times140$) from the League of Robot Runner MAPF competition~\cite{chan2024league}.

In all experiments, the simulation update period is set to 0.1 simulation seconds for both simulators. A simulation update refers to the periodic recalculation of the status of robots and the environment based on the input commands.
In ARGoS3, each grid cell represents 1 meter. Each foot-bot robot is subject to speed limits of [0, 5]~m/s, acceleration limits of [$-$0.4, 0.4]~m/s\textsuperscript{2}, and absolute angular velocity limits of 30$^\circ$/s.
In Isaac Sim, each grid cell represents 0.5 meters. Each JetBot is limited to a speed limit of [0, 2.5]~m/s, an acceleration limit of [$-$0.2, 0.2]~m/s\textsuperscript{2}, and the angular velocity limit of 30$^\circ$/s.

We evaluate SMART using MAPF plans generated by MAPF-LNS2~\cite{li2022mapf}, which provides strong scalability and is representative of state-of-the-art MAPF solvers. Unless otherwise specified, all results in this section are based on MAPF-LNS2.
To verify generality, we additionally conducted experiments using plans generated by other MAPF planners, including CBS~\cite{sharon2015conflict} and PBS~\cite{ma2019searching}.
These plans were created using four different robot models: the standard MAPF model~\cite{Stern2019benchmark}, the $k$-robust delay model~\cite{atzmon2020robust}, MAPF with rotation~\cite{zhang2023efficient}, and MAPF with kinodynamics~\cite{yan2025multi}.
Across all tested settings, SMART achieved a 100\% execution success rate, demonstrating its compatibility with different MAPF planners.

\subsection{Scalability and Runtime Performance} \label{subsec:scale}
This section evaluates the scalability and runtime performance of \mymethod by analyzing its simulation speed, which is defined as the ratio of the simulation time to the real-world time.
A higher simulation speed means that more simulation updates can be finished within the same unit of time, indicating better scalability.
For example, if the simulation speed is 10, the simulation world runs 10 times faster than in real time.
MAPF plans are generated using MAPF-LNS2\footnote{Code at \url{https://github.com/Jiaoyang-Li/MAPF-LNS2}}~\cite{li2022mapf} with a progressively increasing number of robots and a time limit of 60 seconds. For each map, we randomly select 5 scenarios from the MovingAI Benchmark to generate plans, which are then executed in \mymethod without visualization.
In ARGoS3, the maximum number of robots tested corresponds to the limit imposed by the MAPF instances from the MovingAI Benchmark for all maps except \texttt{sortation-center}, which is set to 2,000 robots. In contrast, for Isaac Sim, the upper bound on the number of robots is determined by the point at which the simulation speed drops below real-time (i.e., simulation speed $<$ 1).

\begin{figure}
    \centering
    \includegraphics[width=1.\linewidth]{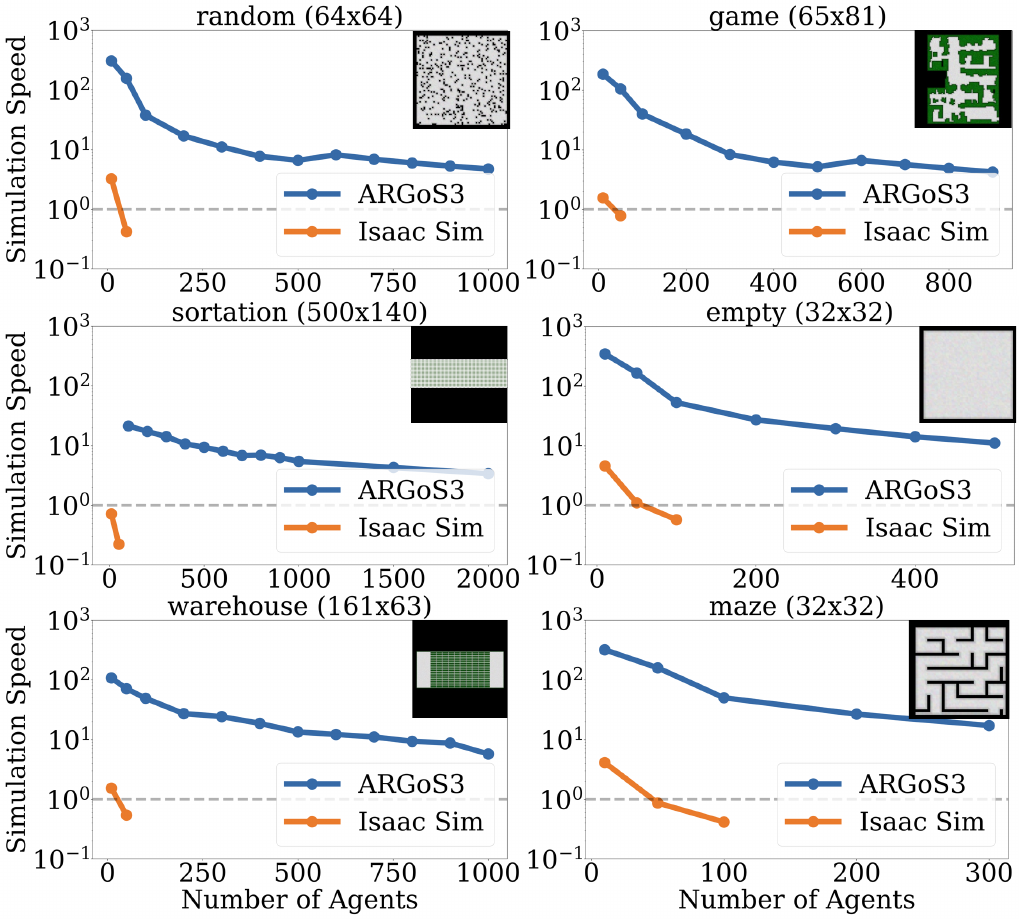}
    \caption{Simulation speed in all environments.}
    \label{fig:runtime-ratio}
\end{figure}

\begin{table}
\small
\centering
\caption{Replicability using different numbers of robots.}
\resizebox{\linewidth}{!}{
\begin{tabular}{crcc}
\toprule
{Simulator} & {Robots} & {average execution} & {maximum execution} \\
& {(\#)} & {time} & {time} \\\midrule
\multirow{5}{*}{{ARGoS3}}& 10 & 123.33 $\pm$ 0.00 & 373.6 $\pm$ 0.00 \\
& 50 & 182.84 $\pm$ 0.01 & 433.30 $\pm$ 0.00 \\
& 100 & 263.36 $\pm$ 0.01 & 469.80 $\pm$ 0.00 \\
& 200 & 269.79 $\pm$ 0.02 & 675.50 $\pm$ 0.00 \\
& 400 & 385.76 $\pm$ 0.02 & 737.60 $\pm$ 0.42 \\
& 600 & 462.87 $\pm$ 0.07 & 843.70 $\pm$ 0.71 \\
& 1000 & 612.04 $\pm$ 0.07 & 1389.43 $\pm$ 0.83 \\
\midrule
\multirow{2}{*}{{Isaac Sim}}& 10 & 122.47 $\pm$ 0.00 & 369.47 $\pm$ 0.00 \\
& 50 & 183.71 $\pm$ 0.07 & 435.46 $\pm$ 1.36 \\
\bottomrule
\end{tabular}}
\label{tab:replicability_summary}
\end{table}

As shown in~\cref{fig:runtime-ratio}, \mymethod supports simulations with up to 2,000 robots. 
As the number of robots increases and the map becomes larger, the simulation speed decreases.
Nevertheless, for ARGoS3, \mymethod maintains a simulation speed of around 10.0 with 1,000 robots in \texttt{warehouse}. Even in the larger \texttt{sortation-center}, the simulation speed is consistently greater than 1.0.
In contrast, Isaac Sim provides the capability to model more complex real-world factors while also producing photorealistic visualizations, but it is significantly less scalable, particularly in large environments. It is therefore more suitable for experiments with a smaller number of robots.

\subsection{Replicability} \label{subsec:repli}
We evaluate the consistency of \mymethod by examining whether the same MAPF plans yield the same outcomes across multiple runs.
Due to the realistic setting we used, uncertainty may arise from factors such as network communication delays or controller variability.
Using the same \texttt{warehouse} map and MAPF plans as in the previous section, we execute \mymethod 10 times for each MAPF plan.
For each run, we record both the average and the maximum execution time.
As shown in~\cref{tab:replicability_summary}, \mymethod demonstrates strong replicability across runs, with minor execution time variations. While these variations increase slightly with more robots, they remain minimal even for up to a thousand robots.

\subsection{CPU and Memory Usage} \label{subsec:usage}
We evaluate the CPU and memory usage of \mymethod on both machines (1) and (2). We use machine (2) to demonstrate that \mymethod can run on resource-constrained machines. To evaluate usage, we generate MAPF plans with MAPF-LNS2 and run \mymethod on the \texttt{warehouse} map with different numbers of robots, each with 5 random MAPF instances selected from the MovingAI Benchmark~\cite{Stern2019benchmark}. All simulations are parallelized on all CPU cores on both machines. We record the peak CPU and memory usage while running the simulation and compute the average of all instances for each number of robots. \Cref{tab:cpu_memory_comparison} shows the results. When using ARGoS3 as the simulator, we observe that \mymethod has reasonable CPU usage and low memory usage on both machines, making it possible to run on low-spec personal laptops. The simulation speed in machine (2) is slower than in machine (1) due to an inferior processor.
In contrast, simulations with Isaac Sim result in significantly higher CPU and memory usage, indicating that Isaac Sim is better suited for scenarios with fewer robots but more powerful hardware.

\begin{table}[t]
\centering
\small
\caption{Comparison of CPU usage, memory consumption, and simulation speed for ARGoS3 and Isaac Sim across two machines and various robot counts. Since Isaac Sim only supports recent GPUs with ray tracing capabilities, we conduct experiments with Isaac Sim only on Machine (1).}
\resizebox{\linewidth}{!}{
\begin{tabular}{c|c|r|ccc}
\toprule
 & {Sim} & {Robots} & {CPU Usage} & {Memory Usage} & {Simulation Speed} \\
 & & {(\#)} & {(\%)} & {(GB)} &  \\
\midrule
\multirow{6}{*}{\rotatebox{90}{Machine (1)}} 
&  \multirow{3}{*}{\rotatebox{90}{ARGoS3}} & 10 &  2.99$\pm$0.10  &  0.03$\pm$0.02  &  107.93$\pm$1.80 \\
& & 50 &  8.03$\pm$0.08  &  0.07$\pm$0.00  &  62.35$\pm$2.33 \\
  &     & 100  & 16.48 $\pm$ 0.58 & 0.19 $\pm$ 0.26 & 33.49 $\pm$ 0.15 \\
  &                                           & 500  & 22.86 $\pm$ 1.59 & 0.24 $\pm$ 0.01 & 13.21 $\pm$ 0.13 \\
  &                                           & 1000 & 46.02 $\pm$ 0.98 & 0.65 $\pm$ 0.03 & 7.37 $\pm$ 0.04 \\
  \cmidrule(r){2-6}
  & \multirow{3}{*}{\rotatebox{90}{Isaac Sim}} & 10 &  25.53$\pm$4.29  &  10.16$\pm$0.82  &  1.35$\pm$0.01 \\
& & 50 &  27.18$\pm$2.97  &  10.94$\pm$7.25  &  0.54$\pm$0.00 \\
  &  & 100  & N/A                & N/A               & N/A \\
  &                                           & 500  & N/A                & N/A               & N/A \\
  &                                           & 1000 & N/A                & N/A               & N/A \\
\midrule
\multirow{3}{*}{\rotatebox{90}{Machine (2)}} 
& \multirow{3}{*}{\rotatebox{90}{ARGoS3}} & 10 &  18.54$\pm$3.88  &  0.01$\pm$0.01  & 37.67 $\pm$ 1.02 \\
& & 50 & 20.20 $\pm$ 0.91 & 0.05 $\pm$ 0.03 & 21.42 $\pm$ 0.35 \\
  &  & 100  & 27.96 $\pm$ 0.39 & 0.04 $\pm$ 0.00 & 14.37 $\pm$ 0.10 \\
  &                                           & 500  & 68.74 $\pm$ 17.24 & 0.22 $\pm$ 0.01 & 3.89 $\pm$ 0.02 \\
  &                                           & 1000 & 59.64 $\pm$ 0.32 & 0.63 $\pm$ 0.02 & 1.95 $\pm$ 0.01 \\
\bottomrule
\end{tabular}
}
\label{tab:cpu_memory_comparison}
\end{table}


\begin{figure}
    \centering
    \includegraphics[width=1.\linewidth]{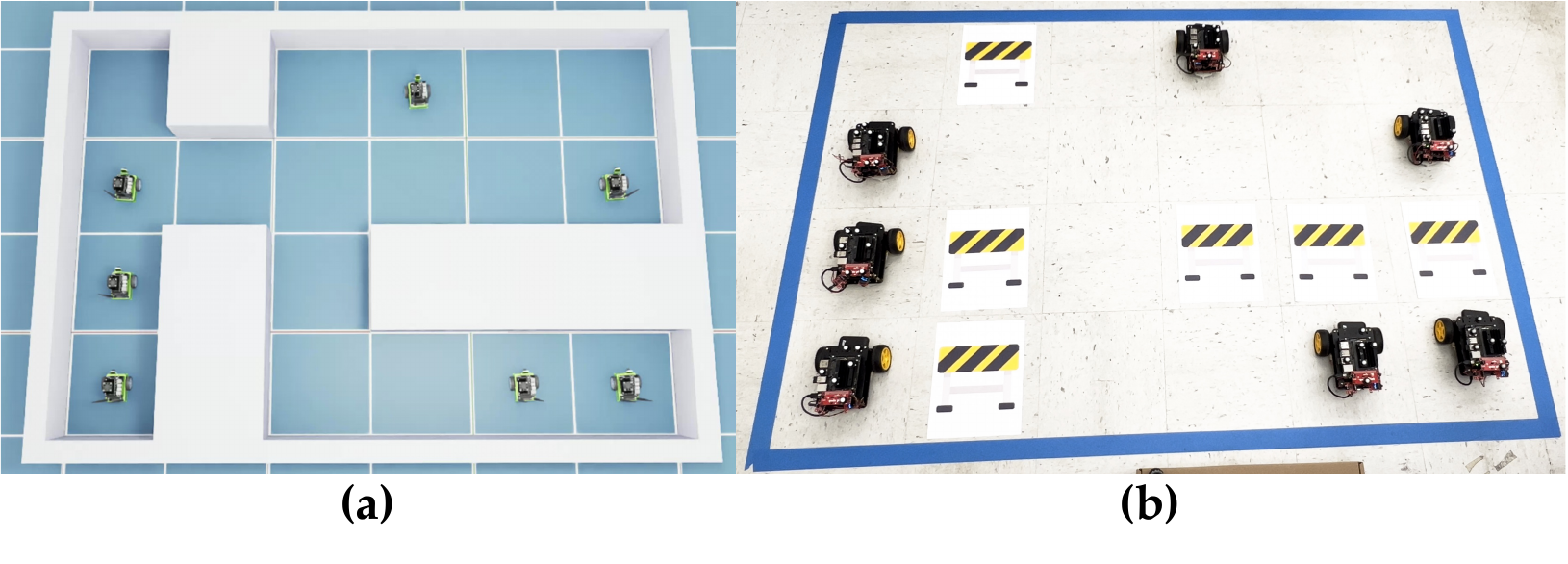}
    \caption{(a) Experimental setup in Isaac Sim. (b) Experimental setup with real robots.}
    \label{fig:realworld-exp}
\end{figure}

\subsection{Evaluation in Real World} 
Finally, we demonstrate the execution of \mymethod on real-world mobile robots. Specifically, we deploy \mymethod on a team of seven JetBot differential-drive robots operating in a structured indoor environment, as illustrated in~\Cref{fig:realworld-exp}.
While laboratory constraints limit our validation to 7 robots, this experiment demonstrates execution robustness under realistic sensing, actuation, and communication conditions rather than physical scalability.
The environment is configured as a 4$\times$6 grid, with each cell measuring 0.3$\times$0.3 meters, and both obstacles and boundaries are physically defined. A Vicon motion capture system provides accurate real-time localization for the robots.

Given a set of start and goal locations, MAPF-LNS2 is used on a central server to generate a MAPF plan. \mymethod then processes this plan and transmits control commands to each robot via a wireless network, with a control frequency of 40 Hz. Each JetBot follows its assigned commands accordingly.

As shown in the accompanying multimedia material, \mymethod robustly executes the MAPF plan on physical robots, highlighting the practical feasibility and reliability of our planning framework in real-world scenarios.
Across 10 trials, \mymethod achieved a 100\% success rate, with an average execution time of 48.48~$\pm$~8.79 seconds and a maximum execution time of 73.83~$\pm$~9.81 seconds.
For reference, in Isaac Sim, the average execution time is 45.66~$\pm$~0.01 seconds, and the maximum execution time is 70.97~$\pm$~0.00 seconds.

\section{Conclusion}
We introduce \mymethod, a scalable testbed for evaluating MAPF planners under realistic settings.
\mymethod bridges the gap between methods that use simplified MAPF models and their real-world deployments.
\mymethod is compatible with existing 2-D grid-based benchmarks, making it straightforward for researchers to evaluate their methods in more realistic environments.
We integrate \mymethod with two simulators, ARGoS3 and Isaac Sim. Our experiments show that \mymethod can handle thousands of robots while maintaining high simulation speeds when using ARGoS3.
SMART's modular, planner-agnostic design supports diverse application-driven use cases across academic and industrial settings. Academic researchers can easily benchmark different algorithms without custom integration, while industrial users can evaluate algorithms in their specific warehouse or manufacturing environments without requiring specialized MAPF expertise.
Future work includes: (1) extending to lifelong MAPF~\cite{ma2017lifelong} where tasks arrive continuously and require dynamic replanning, (2) integrating sophisticated dynamic obstacle models for human-robot interaction scenarios, (3) supporting alternative execution frameworks~\cite{coskun2019repairMAPF,berndt2023receding,feng2024real}.


\section*{Acknowledgments}
The research was supported by the National Science Foundation (NSF) under grant numbers \#2328671 and \#2441629, as well as a gift from Amazon.
The views and conclusions contained in this document are those of the authors and should not be interpreted as representing the official policies, either expressed or implied, of the sponsoring organizations, agencies, or the U.S. government.



\bibliographystyle{IEEEtran}
\bibliography{ral}


 





\end{document}